\documentclass[11pt,a4paper]{article}
\usepackage[hyperref]{emnlp-ijcnlp-2019}
\usepackage{times}
\usepackage{latexsym}
\usepackage{xspace}
\usepackage{enumitem}

\usepackage{url}

\usepackage{booktabs} 
\usepackage{diagbox}

\usepackage{multirow}
\usepackage{balance}
\usepackage{amsfonts}
\usepackage{amsmath}
\usepackage{amssymb}
\usepackage{graphicx}
\usepackage{subfigure}
\usepackage{microtype}
\usepackage{xspace}
\usepackage{wrapfig,lipsum}
\usepackage[noend,ruled]{algorithm2e}
\usepackage{makecell}
\usepackage[T1]{fontenc}

\usepackage{tabularx, booktabs}
\newcolumntype{Y}{>{\centering\arraybackslash}X}
\newcolumntype{L}{>{\arraybackslash}X}
\aclfinalcopy 

\newcommand{\our}{\mbox{\sf CrossWeigh}\xspace}

\newcommand{\smallsection}[1]{{\noindent\textbf{#1.}}}

\usepackage[normalem]{ulem}

\title{CrossWeigh: Training Named Entity Tagger from Imperfect Annotations}

\author{
Zihan Wang\thanks{\; Equal Contributions.}$\quad$
Jingbo Shang$^{*}\ $
Liyuan Liu$^{*}\ $
Lihao Lu$\quad$ 
Jiacheng Liu$\quad$ 
Jiawei Han
\\[0.5ex]
University of Illinois at Urbana-Champaign, Urbana, IL, USA\\
\{zihanw2, shang7, ll2, lihaolu2, jl25, hanj\}@illinois.edu
}

\date{}

\begin{document}
\maketitle
\begin{abstract}
  Everyone makes mistakes.
So do human annotators when curating labels for named entity recognition (NER). 
Such label mistakes might hurt model training and interfere model comparison.
In this study, we dive deep into one of the widely-adopted NER benchmark datasets, CoNLL03 NER.
We are able to identify label mistakes in about 5.38\% test sentences, which is a significant ratio considering that the state-of-the-art test F$_1$ score is already around 93\%.
Therefore, we manually correct these label mistakes and form a cleaner test set.
Our re-evaluation of popular models on this corrected test set leads to more accurate assessments, compared to those on the original test set.
More importantly, we propose a simple yet effective framework, \our, to handle label mistakes during NER model training.
Specifically, it partitions the training data into several folds and train independent NER models to identify potential mistakes in each fold.
Then it adjusts the weights of training data accordingly to train the final NER model.
Extensive experiments demonstrate significant improvements of plugging various NER models into our proposed framework on three datasets. 
All implementations and corrected test set are available at our Github repo \footnote{\url{https://github.com/ZihanWangKi/CrossWeigh}}.

\end{abstract}

\section{Introduction}

Named entity recognition (NER), identifying both spans and types of named entities in text, is a fundamental task in the natural language processing pipeline.
On one of the widely-adopted NER benchmarks, the CoNLL03 NER dataset~\cite{sang2003introduction}, the state-of-the-art NER performance has been pushed to a F$_1$ score around 93\%~\cite{akbikpooled},
through building end-to-end neural models~\cite{lample2016neural,ma2016end} and introducing language models for contextualized representations~\cite{peters2017semi,peters2018deep,akbik2018contextual,liu2018efficient}.
Such high performance makes the label mistakes in manually curated ``gold standard'' data non-negligible.
For example, given a sentence ``\emph{Chicago won game 1 with Derrick Rose scoring 25 points.}'', this ``\emph{Chicago}'', representing the NBA team Chicago Bulls, should be annotated as an organization.
However, when annotators are not careful or lack background knowledge, this ``\emph{Chicago}'' might be annotated as a location, thus being a label mistake.

These label mistakes bring up two challenges to NER:
(1) mistakes in the test set can interfere the evaluation results and even lead to an inaccurate assessment of model performance;
and (2) mistakes in the training set can hurt NER model training.
Therefore, in this paper, we conduct empirical studies to understand these mistakes, correct the mistakes in the test set to form a cleaner benchmark, and develop a novel framework to handle the mistakes in the training set.

We dive deep into the CoNLL03 NER dataset, and find label mistakes in about 5.38\% test sentences.
Considering that the state-of-the-art F$_1$ score on this test set is already around 93\%, these 5.38\% mistakes should be considered as significant.
So we hire human experts to correct these label mistakes in the test set.
We then re-evaluate recent state-of-the-art NER models on this new, cleaner test set.
Compared to the results on the original test set, the re-evaluation results are more accurate and stable.
Therefore, we believe this new test set can better reflect the performance of NER models.

\begin{figure*}[t]
    \centering
    \includegraphics[width=1\linewidth]{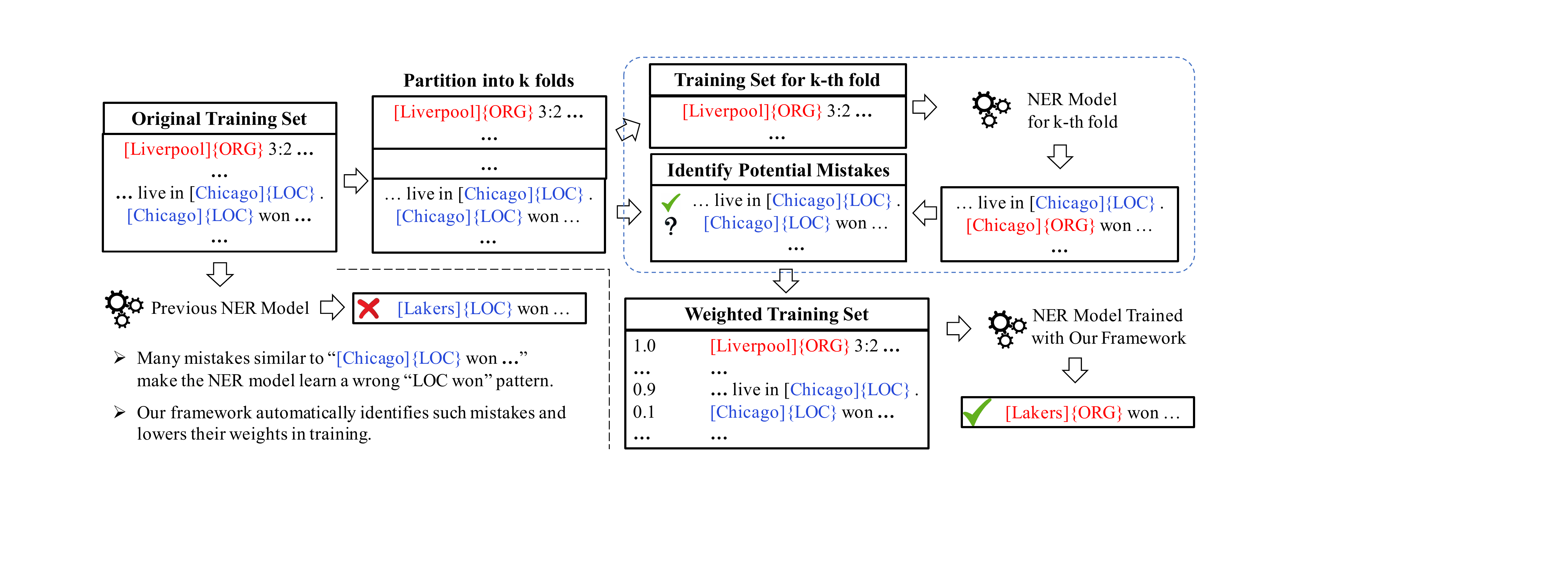}
    \caption{An overview of our proposed \our framework. 
    It can better handle label mistakes, identify low quality annotations and conduct learning from a weighted training set. }
    \label{fig:framework}
\end{figure*}

We further propose a novel, general framework, \our, to handle the label mistakes during the NER model training stage. 
Figure~\ref{fig:framework} presents an overview of our proposed framework.
It contains two modules: (1) mistake estimation: it identifies the potential label mistakes in training data through a cross checking process and (2) mistake re-weighing: it lowers the weights of these instances during the training of the final NER model.
The cross checking process is inspired by the k-fold cross validation; differently, in each fold's training data, it removes the data containing any of entities that appeared in this fold.
In this way, each sentence will be scored by a NER model trained on a subset of training data not containing any entity in this sentence.
Once we know where the potential mistakes are, we lower the weights of these sentences and train the final NER model based on this weighted training set.
The final NER model is trained in a mistake-aware way, thus being more accurate.
Note that, our proposed framework is general and fits most of, if not all, NER models that accept weighted training data.

To the best of our knowledge, we are the first to handle
the label mistake issues systematically in the NER problem. 
We conduct extensive experiments on both the original CoNLL03 NER dataset and our corrected dataset. 
\our is able to consistently improve performance when plugging with different NER models.
In addition, we verify the effectiveness of \our on emerging-entity and low-resource NER datasets.
In summary, our major contributions are the following:
\begin{itemize}[leftmargin=*,nosep]
    \item We correct label mistakes in the test set of the CoNLL03 NER dataset and re-evaluate popular NER models.
    This establishes a more accurate NER benchmark.
    \item We propose a novel framework \our to accommodate the mistakes during the model training stage. 
    The proposed framework fits most of, if not all, NER models.
    \item Extensive experiments demonstrate the significant, robust test F$_1$ score improvements of plugging NER models into our proposed framework on three datasets, not only CoNLL03 but also emerging-entity and low-resource datasets.
\end{itemize}

\smallsection{Reproducibility}
We release both the corrected test set and the implementation of \our framework\footnote{\url{https://github.com/ZihanWangKi/CrossWeigh}}.


\begin{table*}[t]
\centering
\vspace{-0.3cm}
\small
\scalebox{0.9}{%
\begin{tabularx}{1.11\linewidth}{lll}
\toprule
\textbf{Sentence} & \textbf{Original labels} & \textbf{Corrected labels} \\
\midrule
Sporting Gijon 15 4 4 7 15 22 16 & [Sporting]\{ORG\} & [Sporting Gijon]\{ORG\} \\
\midrule
SOCCER - JAPAN GET LUCKY WIN , & [JAPAN ]\{LOC\}, [China]\{PER\} & [JAPAN]\{LOC\}, [China]\{LOC\}\\
CHINA IN SURPRISE DEFEAT .  & & \\
\midrule
NZ 's Bolger says Nats to meet & [NZ]\{LOC\}, [Bolger]\{PER\},  & [NZ]\{LOC\}, [Bolger]\{PER\},  \\
NZ First on Sunday .  & [Nats]\{PER\}, [NZ]\{LOC\} & [Nats]\{ORG\}, [NZ First]\{ORG\} \\
\midrule
Seagramd ace 20/11/96 5,000 Japan & [Seagramd] \{MISC\}, [Japan]\{LOC\} & [Seagramd ace]\{MISC\}, [Japan] \{LOC\}\\
\bottomrule
\end{tabularx}
}
\vspace{-0.15cm}
\caption{Typical Examples of Our Corrections on the CoNLL03 NER dataset.}
\label{tbl:annotation_mistake}
\vspace{-0.4cm}
\end{table*}

\section{CoNLL03 NER Re-Examination}\label{sec:correction}
    
    The CoNLL03 NER dataset is one of the widely-adopted NER benchmark datasets. 
    Its annotation guideline is based on MUC Conventions\footnote{\url{https://www-nlpir.nist.gov/related_projects/muc/proceedings/ne_task.html}}~\cite{sang2003introduction}.
    Following this guideline, the annotators are asked to mark entities of person~(\texttt{PER}), location~(\texttt{LOC}), and organization~(\texttt{ORG}), while using an extra miscellaneous~(\texttt{MISC}) type to deal with entities that do not fall in these categories.
    This dataset has been split into training, development, and test sets, with $14041$, $3250$, and $3453$ sentences, respectively.

    \subsection{Test Set Correction}
        In order to understand and correct the label mistakes, we have hired 5 human experts as annotators.
        Before looking at the data, we first train the annotators by carefully going through the aforementioned guideline.
        During the correction process, we strongly encourage the annotators to use search engines for suspicious token spans.
        This helps them have more background knowledge.
        We also allow annotators to look at the original paragraph containing the sentence.
        This helps them have a better understanding of the context.
        
        For the whole test set, we randomly split the test sentences between each pair combination of 5 annotators. 
        In this way, each sentence in the test set is checked by exactly two annotators. 
        The inter-annotator agreement is 95.66\%.
        This is a reasonable score, given that the inter-annotator agreement in POS tagging annotations is about 97\%~\cite{manning2011part}.
        After we collected all annotations, we run a final round of verification on each sentence, where the original annotation and the two annotators' are not all the same.
        In the end, we have corrected label mistakes in 186 sentences, which is about $5.38\%$ of the test set.
        
        Table~\ref{tbl:annotation_mistake} presents some typical examples of our corrections.
        In the first sentence, as a sport team, ``Sporting Gijon'' was not annotated completely. 
        In the second sentence, while ``JAPAN'' is correctly marked as \texttt{LOC},  ``China'' is wrongly identified as \texttt{PER} instead of \texttt{LOC}. 
        One may notice that they both represent sport teams.
        However, according to the aforementioned guideline, country names should be marked as \texttt{LOC} even when they are sports teams. 
        More details about this type of labels are discussed in Section~\ref{sec:case}. 
        In the third sentence, ``NZ'' is the abbreviation of New Zealand.
        However, ``Nat'' and ``NZ First'' in fact refer to political parties (i.e., New Zealand Young Nationals and New Zealand First). 
        So they should be labelled as \texttt{ORG}. 
        In the forth sentence, looking at its paragraph, our annotators figure out that this is a table about ships and vessels loading items at different locations. 
        Through comparing with other sentences in the context, such as ``Algoa Day	21/11/96 6,000 Africa'', our annotators identified ``Seagramd ace'' as a vessel, thus marking it as \texttt{MISC}.
        We have verified that there is indeed a vessel called ``Seagrand Ace'' (``Seagramd ace'' might be a typo). 
    
    \subsection{CoNLL03 Re-Evaluation} \label{sec:re-eval}
    
        \smallsection{NER Algorithms}
        We re-evaluate following popular NER algorithms:
        \begin{itemize}[leftmargin=*,nosep]
            \item \textbf{LSTM-CRF}~\cite{lample2016neural} incorporates long short term memory (LSTM) neural network with conditional random field (CRF). It also uses a word-wise character LSTM.
            \item \textbf{LSTM-CNNs-CRF}~\cite{ma2016end} has a similar structure as LSTM-CRF, but captures character-level information through a convolutional neural network (CNN) over the character embedding.
            \item \textbf{VanillaNER}~\cite{liu2018efficient} also extends LSTM-CRF and LSTM-CNNs-CRF by using a sentence-wise character LSTM. 
            \item \textbf{ELMo}~\cite{peters2018deep} extends LSTM-CRF and leverages pre-trained word-level language models for better contextualized representations.
            \item \textbf{Flair}~\cite{akbik2018contextual} also aims for contextualized representations, utilizing pretrained character level language models.
            \item \textbf{Pooled-Flair}~\cite{akbik2018contextual} extends Flair and maintains an embedding pool for each word to bring in dataset-level word embedding.
        \end{itemize}
        We use the implementation released  by the authors for each algorithm and report the performance on original test set and corrected test set averaging 5 runs.
        
\begin{table}[t]
\begin{center}
\small
\begin{tabularx}{\linewidth}{l *{2}{Y}}
\toprule
       \textbf{Method}   &  \textbf{Original} & \textbf{Corrected} \\ \midrule
LSTM-CRF & 90.64 ($\pm$0.23)& 91.47 ($\pm$0.15)\\
LSTM-CNNs-CRF & 90.65 ($\pm$0.57) & 91.87 ($\pm$0.50)\\
VanillaNER & 91.44 ($\pm$0.16) & 92.32 ($\pm$0.16) \\
Elmo & 92.28 ($\pm$0.19) & 93.42 ($\pm$0.15) \\
Flair & 92.87 ($\pm$0.08) & 93.89 ($\pm$0.06) \\
Pooled Flair & 93.14 ($\pm$0.14) &  94.13 ($\pm$0.11) \\
\bottomrule
\end{tabularx}
\end{center}
\vspace{-0.15cm}
\caption{CoNLL03 Re-Evaluation: Test F$_1$ scores and standard deviations on both original and corrected datasets. The results are based on $5$ different runs.}
\label{tbl:reevaluation}
\vspace{-0.4cm}
\end{table}

        
        \smallsection{Results \& Discussions}
            We re-evaluate the performance of the NER algorithms on the corrected test set.
            Their performance on the original test set is also listed for the reference.
            From the results in Table~\ref{tbl:reevaluation}, one can observe that all models have higher F$_1$ scores as well as smaller standard deviations on the corrected test set, compared to those on the original test set.
            Moreover, LSTM-CRF has a similar performance as LSTM-CNNs-CRF on the original test set, but on average lower performance on the corrected test set. This indicates that the corrected test set may be more discriminative.
            Therefore, we believe this corrected test set can better reflect the accuracy of NER algorithms in a stable way.

\section{Our Framework: CrossWeigh}\label{sec:crossweigh}
    In this section, we introduce our framework.
    It is worth mentioning that our framework is designed to be general and fits most of, if not all, NER models.
    The only requirement is the capability to consume weighted training set.
    
    \subsection{Overview}
    
        As we have seen in the Section~\ref{sec:correction}, human curated NER datasets are by no means perfect.
        Label mistakes in the training set can directly hurt the model's performance. 
        As shown in Figure~\ref{fig:framework}, if there are many similar mistakes like wrongly annotating ``Chicago'' in ``Chicago won ...'' as \texttt{LOC} instead of \texttt{ORG}, the NER model will likely capture the wrong pattern ``\texttt{LOC} won'' and make wrong predictions in future.
        
        
        Our proposed \our framework automates this process.
        Figure~\ref{fig:framework} presents an overview.
        It contains two modules: (1) mistake estimation: it identifies the potential label mistakes in training data through a cross checking process and (2) mistake re-weighing: it lowers the weights of these instances for the NER model training. The workflow is summarized in Algorithm~\ref{alg:model_algo}.
        
\SetAlgoSkip{}
\begin{algorithm}[t]
    \caption{Our CrossWeigh Framework}\label{alg:model_algo}
    \textbf{Input}: A NER model $f$, the training set $\mathbf{D}$ = $\langle \{x_1, \ldots, x_n\}, \{y_1, \ldots, y_n\} \rangle$, and hyper-parameters $k$, $t$, and $\epsilon$. \\
    \textbf{Output}: A final NER model \\
    \For{i = 1 ... n} {
        $c_i \leftarrow 0, w_i \leftarrow 1$
    }
    \For{iter = $1 \ldots t$} {
        Randomly partition $\mathbf{D}$ into $k$ folds. \\
        \For{Each fold $D_i$} {
            Obtain $\mbox{test\_entities}_i$. (Eq.~\ref{eq:test_entities})\\
            Build $\mbox{train\_set}_i$. (Eq.~\ref{eq:train_set}).\\
            Train a NER model $M_i = f(\mbox{train\_set}_i, w)$.\\
            \For{Each $x_j \in D_i$} {
                $\hat{y}_j \leftarrow$ $M_i$'s prediction on $x_j$.\\
                \If{$y_j \neq \hat{y}_j$}{
                    $c_i \leftarrow c_i + 1$
                }
            }
        }
    }
    \For{i = $1 \ldots n$} {
        Compute $w_i$ (Eq.~\ref{eq:weight}). \\
    }
    \textbf{Return} $f(\mathbf{D}, w)$.\\
\end{algorithm}

    \subsection{Preliminary}
        We denote the training sentences as $\{x_1, x_2, \ldots, x_n\}$ where $n$ is the number of sentences. 
        Each sentence $x_i$ is formed up of a sequence of words.
        Correspondingly, the label sequence for each sentence is denoted as $\{y_1, y_2, \ldots, y_n\}$.
        We use $\mathbf{D}$ to denote the training set, including both sentences and their labels.
        We use $w_i$ to represent the weight of the $i$-th sentence.
        In most NER papers, the weights are uniform, i.e., $w_i = 1$.
        
        We use $f(\mathbf{D}, w)$ to describe the training process of an NER model using the training set $\mathbf{D}$ weighted by $w$. 
        This training process will return an NER model $M = f(\mathbf{D}, w)$.
        During this training, the weighted loss function is as below.
        \begin{equation}
            \mathcal{J} = \sum_{i = 1}^{n} w_i \cdot l(M(x_i), y_i)
        \end{equation}
        where $l(M(x_i), y_i)$ is the loss function of prediction $M(x_i)$ against its label sequence $y_i$.
        Typically, it is the negative log-likelihood of the model's prediction $M(x_i)$ compared to labeling sequence $y_i$.
        
\begin{table*}[t]
\center
\small
\begin{tabularx}{0.85\linewidth}{l *{4}{Y}}
\toprule
                & \multicolumn{2}{c}{\textbf{Original CoNLL03}} & \multicolumn{2}{c}{\textbf{Corrected CoNLL03}}\\
\cmidrule{2-5}
                & w/o \our  & w/ \our  & w/o \our  & w/ \our \\
\midrule
VanillaNER & 91.44 ($\pm$0.16) & \textbf{91.78 ($\pm$0.06)} & 92.32 ($\pm$0.16) & \textbf{92.64 ($\pm$0.08)} \\
Flair  & 92.87 (\textbf{$\pm$0.08}) & \textbf{93.19} ($\pm$0.09) & 93.89 (\textbf{$\pm$0.06}) & \textbf{94.18 ($\pm$0.06)} \\
Pooled-Flair  & 93.14 ($\pm$0.14) & \textbf{93.43 ($\pm$0.06)} & 94.13 ($\pm$0.11) & \textbf{94.28 ($\pm$0.05)} \\
\bottomrule
\end{tabularx}
\vspace{-0.15cm}
\caption{Test F$_1$ scores and its standard deviations of models trained without or with \our.}
\label{tbl:performance}
\vspace{-0.4cm}
\end{table*}



    \subsection{Mistake Estimation}
        Our mistake estimation module is designed to let an NER model itself decide which sentences contain mistake and which do not. 
        We would like to find sentences with label mistakes as many as possible (i.e. high recall), while keeping away from wrongly identified non-mistake sentences (i.e. high precision).
        
        The basic idea of our mistake estimation module is similar to k-fold cross validation, however, in each fold's training data, it further removes the data containing any of entities appearing in this fold. 
        The details are presented as follows.
        
        We first randomly partition the training data into $k$ folds: $D_1, D_2, \ldots, D_k$.
        
        We then train $k$ NER models separately based on these $k$ folds.
        The $i$-th ($1 \le i \le k$) NER model $M_i$ will be evaluated on the sentences in the hold-out fold $D_i$. 
        
        During its training, we avoid any sentence that may lead to ``easy prediction'' on this hold-out set. 
        Therefore, we inspect every sentence in $D_i$ and get the set of entities as follows.
        \begin{equation}
        \label{eq:test_entities}
            \mbox{test\_entities}_i = \bigcup_{x_j \in D_i} e_{j}
        \end{equation}
        where $e_j$ is the set of named entities in sentence $x_j$.
        We only consider the surface name in this entity set. 
        That is, no matter ``Chicago'' is \texttt{LOC} or \texttt{ORG}, it only counts as its surface name ``Chicago''.
        
        All training sentences that have entities included in $\mbox{test\_entities}_i$ will be excluded in training process of the model $M_i$.
        Specifically,
        \begin{equation}
        \label{eq:train_set}
            \mbox{train\_set}_i = \{ x_j | e_j \cap \mbox{test\_entities}_i = \emptyset, 1 \leq j \leq n \}
        \end{equation}
        We call this step as \textit{entity disjoint filtering}. 
        The intuition behind this step is that we want the model to make prediction of an entity without prior information of the entity itself from training. 
        This will be helpful to detect sentences that are inconsistent.
        
        We train $k$ models $M_i$ by feeding each $\mbox{train\_set}_i$ into $f(\cdot, \cdot)$ with default uniform weight, and we use each $M_i$ to make predictions for $D_i$ and check for each sentence, whether the original label is the same as the model output. 
        In this way, if the trained model $M_i$ makes correct predictions on some sentences in $D_i$, they are more likely mistake-free.
        For those sentences that have labels disagreeing with the model output, we mark them as \textit{potentially mistake}. 
        
        We run this mistake estimation module multiple iterations (i.e. $t$ iterations) using different random partitions.
        Then, for each sentence in the training set, we get $t$ estimations for it. 
        We denote $c_i$ ($0 \le c_i \le t$) as the confidence that sentence $x_i$ contains label mistakes.
        $c_i$ is defined as the the number of \textit{potentially mistake} indications among all $t$ estimations.

        The number of folds $k$ plays the role of a trade-off between the efficiency of the mistake estimation process and the number of training examples that can be used in each $M_i$.
        When $k$ becomes larger, each fold $D_i$ will be smaller, thus leading to a smaller size of $\mbox{test\_entities}_i$;
        correspondingly, a larger $\mbox{train\_set}_i$ will be picked.
        The model can therefore be trained with more examples. 
        However, it also slows down the whole mistake estimation process.
        On the CoNLL03 NER dataset, we observe that $k = 10$ leads to effective results, while having a reasonable running time.

    \subsection{Mistake Reweighing} 
    
        In the mistake reweighing module, we adjust weight $w_i$ for each sentence $x_i$ that is marked as \textit{potentially mistake} in the mistake estimation step.
        Here, we assign a weight $w_i$ to all sentences marked, while the weights of other sentences remain $1$. 
        Specifically, we set $\forall i (1 \le i \le n)$, 
        \begin{equation}\label{eq:weight}
            w_i = \epsilon^{c_i}
        \end{equation}
        where $\epsilon$ is a parameter.
        In practice, it can be chosen according to the quality of mistake estimation module. 
        Particularly, we first estimate the precision of the detected mistakes of a single iteration. 
        Let $p$ be the ratio of the number of true detected label mistakes over the number of detected label mistakes.
        $p$ can be roughly estimated through a manual check of a random sample from the detected label mistakes.
        Then, we choose $\epsilon = 1 - p$, because $1 - p$ represents the fraction of these detected label mistakes that might be still useful during the model training. 
        Therefore, for the sentences that are marked as \textit{potentially mistake} in that iteration, $\epsilon$ of them are actually correct. 
        With more iterations, the confidence of being correct lowers like a binomial distribution, which is the reason that we chose an exponential decaying weight function in Equation~\ref{eq:weight}. 

\section{Experiments}\label{sec:exp}

In this section, we conduct several experiments to show effectiveness of our \our framework.
We first evaluate the overall performance of \our on benchmark NER datasets, by plugging it into three base NER models.
Since we have two modules in \our, we then dive into each module and explore different variants and ablations.
In addition, we further verify the effectiveness of \our on two more datasets: an emerging-entity NER dataset from WNUT'17 and a low-resource language NER dataset of the Sinhalese language.

    \subsection{Experimental Settings}
    
        \smallsection{Dataset}
        We use both the original and corrected CoNLL03 datasets.
        We follow the standard train/dev/test splits and use both the train set and dev set for training ~\cite{peters2017semi, akbik2018contextual}.
        Entity-wise F$_1$ score on the test set is the evaluation metric.
    
        \smallsection{Base NER Algorithm}
        We mainly choose Flair as our base NER algorithm. Flair is a strong NER algorithm using external resources (large corpus to train a language model). While Pooled-Flair has even better performance, its computational cost refrains us from doing extensive experiments. 

        \smallsection{Default Parameters in \our}
        For all NER algorithms we experiment with, their default parameters are used.  
        For \our parameters, by default, we set $k = 10$, $t = 3$, and $\epsilon = 0.7$. 
        We decide $\epsilon = 0.7$ because among $100$ randomly sampled sentences with \textit{potentially mistake}, we find that $27$ of them really contain label mistakes (i.e., the probability of one annotation to be correct is roughly $70\%$). We use both train and development set to train the models, and report average F$_1$ and its standard deviation on both original test set and our corrected test set across 5 different runs~\cite{peters2017semi}. 
    
    \subsection{Overall Performance}
        
        We pair \our with our base algorithm (i.e. Flair) and two best-performing NER algorithms with or without language models in Table~\ref{tbl:reevaluation} (i.e. Pooled-Flair and VanillaNER), and evaluate their performance.
        As shown in Table~\ref{tbl:performance}, compared with the three algorithms, applying \our always leads to a higher F$_1$ score and a comparable, sometimes even smaller, standard deviation.
        Therefore, it is clear that \our can improve the performance of NER models. 
        The smaller standard deviations also imply that the models trained with \our are more stable.
        All these results illustrate the superiority of training with \our.
        
\begin{table}[t]
\center
\small
\begin{tabularx}{1\linewidth}{l *{2}{Y}}
\toprule
\textbf{Method} & \textbf{Original} & \textbf{Corrected}\\
\midrule
w/o \our & 92.87 ($\pm$0.08) & 93.89 ($\pm$0.06)  \\
\midrule
w/ \our & 93.19 ($\pm$0.09) & 94.18 ($\pm$0.06) \\
$\quad$ $-$ Entity Disjoint & 92.88 ($\pm$0.11) & 93.84 ($\pm$0.08) \\
$\quad$ $+$ Random Discard & 93.01 ($\pm$0.10) & 93.94 ($\pm$0.10)\\
\bottomrule
\end{tabularx}
\vspace{-0.15cm}
\caption{Importance of Entity Disjoint Filtering.}
\label{tbl:entity_disjoint}
\vspace{-0.3cm}
\end{table}


%

    \subsection{Ablations and Variants}
        We pick Flair as the base algorithm to conduct ablation study.
        
        \smallsection{Entity Disjoint Filtering}
        There is an entity disjoint filtering step, when we are collecting training data $\mbox{train\_set}_i$ for the NER model $M_i$ during the mistake estimation step. 
        To study its importance, we have done a few ablation experiments. 
        
        We have evaluated the following variants:
        \begin{itemize}[leftmargin=*,nosep]
            \item Flair w/ \our \textbf{-- Entity Disjoint}: Skip the entity disjoint filtering step.
            \item Flair w/ \our \textbf{+ Random Discard}:
            Instead of entity disjoint filtering, randomly discard the same number of sentences from each $\mbox{train\_set}_i$ as it would do.
        \end{itemize}
        
        The results are listed in Table~\ref{tbl:entity_disjoint}.
        One can easily observe that without the entity disjoint filtering, the F$_1$ scores are very close to the raw Flair model. 
        This demonstrates that the entity disjoint filtering is critical to reduce the over-fitting risk in the mistake estimation step.
        Also, our proposed entity disjoint filtering strategy works more effective than random discard. 
        This further confirms the effectiveness of entity disjoint filtering.
    
        \smallsection{Variants in Computing $c_i$}
        There is definitely more than one way to determine $c_i$. Let $\delta$ be the number of ``potentially mistake''s among the $t$ estimations, we can apply any of the following heuristics:
        \begin{itemize}[leftmargin=*,nosep]
            \item \textbf{Ratio}: $c_i$ is the number of  ``potentially mistake'' (i.e. $c_i = \delta$). This is the method mentioned in Section~\ref{sec:crossweigh}, and used by default.
            \item \textbf{At Least One}: $c_i$ is the indicator of at least one estimation being ``potentially mistake'' (i.e. $c_i = t \iff \delta >= 1$).
            \item \textbf{Majority}: $c_i$ is the indicator of at least $\lfloor t/2 \rfloor + 1$ estimations being ``potentially mistake''
            (i.e. $c_i = t \iff \delta >= \lfloor t/2 \rfloor + 1$).
            \item \textbf{All}: $c_i$ is the indicator of all $t$ estimations being ``potentially mistake'' (i.e. $c_i = t \iff \delta = t$).
        \end{itemize}
        We evaluate the performance of these heuristics when used in \our, as shown in Table~\ref{tbl:different_c_i}.
        There is not much difference across these heuristics, while our default choice ``Ratio'' is the most stable.

    \subsection{Label Mistake Identification Results}

        Another usage of \our is to identify potential label mistakes during label annotation process, thus improving the annotation quality.
        This could be also helpful to active learning.
        
        Specifically in this experiment, we apply our noise estimation module to the concatenation of training and testing data. 
        As we have manually corrected the label mistakes in the testing set, we are able to report the number of true mistakes among the potential mistakes discovered in the test set.
        
        The results are presented in  Table~\ref{tbl:noise_estimation}. 
        The potential mistakes are the total number of mistakes identified by \our, and actual mistakes is the true positives among all identifications.
        From the results, we can see that when the base model is Flair, \our is able to spot more than 75\% of label mistakes, 
        while maintaining a precision about 25\%.
        It is worth noting that 25\% is a reasonably high precision, given that the label mistake ratio is only 5.38\%.
        The 75\% recall indicates that \our is able to identify most of the label mistakes, which are extremely valuable to improve the annotation quality.


\begin{table}[t]
\center
\small
\begin{tabularx}{\linewidth}{l *{2}{Y}}
\toprule
\textbf{Heuristic} & \textbf{Original} & \textbf{Corrected}\\

\midrule
At Least One & 93.10 ($\pm$0.10) & 94.16 ($\pm$0.07) \\
Majority & 93.20 ($\pm$0.09) & 94.12 ($\pm$0.07) \\
All & 93.16 ($\pm$0.09) & 94.11 ($\pm$0.09) \\
\midrule
Ratio & 93.19 ($\pm$0.09) & 94.18 ($\pm$0.06) \\
\bottomrule
\end{tabularx}
\vspace{-0.15cm}
\caption{Different $c_i$ Estimation Heuristics.}
\label{tbl:different_c_i}
\vspace{-0.3cm}
\end{table}

\begin{table}[t]
\center
\small
\begin{tabularx}{1\linewidth}{l *{2}{Y}}
\toprule   $t$        & \textbf{Original} & \textbf{Corrected}\\

\midrule
 1 & 93.09 ($\pm$0.14) & 94.07 ($\pm$0.09)  \\
 5 & 93.23 ($\pm$0.10) & 94.14 ($\pm$0.08)  \\
 \midrule
 3 & 93.19 ($\pm$0.09) & 94.18 ($\pm$0.06)  \\

\bottomrule
\end{tabularx}
\vspace{-0.15cm}
\caption{Different Numbers of Iterations $t$.}
\label{tbl:noise_t}
\vspace{-0.3cm}
\end{table}

\begin{table}[t]
\center
\small
\begin{tabularx}{1\linewidth}{l *{2}{Y}}
\toprule  $k$        & \textbf{Original} & \textbf{Corrected}\\

\midrule
 2 & 92.11 ($\pm$0.24) & 92.88 ($\pm$0.11)  \\
 5 & 93.12 ($\pm$0.08) & 94.12 ($\pm$0.08)  \\
 \midrule
 10 & 93.19 ($\pm$0.09) & 94.18 ($\pm$0.06)  \\

\bottomrule
\end{tabularx}
\vspace{-0.15cm}
\caption{Different Numbers of Folds $k$.}
\label{tbl:noise_k}
\vspace{-0.3cm}
\end{table}

\begin{table}[t]
\center
\small
\begin{tabularx}{1\linewidth}{l *{2}{Y}}
\toprule   $\epsilon$        & \textbf{Original} & \textbf{Corrected}\\

\midrule
 0.3 & 92.79 ($\pm$0.14) & 93.68 ($\pm$0.15)  \\
 0.5 & 93.21 ($\pm$0.09) & 94.18 ($\pm$0.07)  \\
 0.9 & 93.01 ($\pm$0.10) & 93.96 ($\pm$0.09)  \\
 \midrule
 0.7 & 93.19 ($\pm$0.09) & 94.18 ($\pm$0.06)  \\

\bottomrule
\end{tabularx}
\vspace{-0.15cm}
\caption{Different Weight Adjustments $\epsilon$.}
\label{tbl:noise_reweigh}
\vspace{-0.3cm}
\end{table}

\begin{table*}[t]
\begin{center}
\vspace{-0.3cm}
\scalebox{0.77}{
\begin{tabularx}{1.27\linewidth}{l *{5}{Y}}
\toprule
 Algorithm & 
\begin{tabular}[c]{@{}l@{}}
Potential Mistakes
\end{tabular} & 
\begin{tabular}[c]{@{}l@{}}
Actual Mistakes
\end{tabular} & 
\begin{tabular}[c]{@{}l@{}}
Precision
\end{tabular} & 
\begin{tabular}[c]{@{}l@{}}
Recall
\end{tabular} &
\begin{tabular}[c]{@{}l@{}}
F1
\end{tabular}
\\\midrule

\begin{tabular}[c]{@{}c@{}}
\textbf{Flair}
\end{tabular}
& 
\begin{tabular}[c]{@{}l@{}}
573.0
\end{tabular}
& 
\begin{tabular}[c]{@{}l@{}}
144.0
\end{tabular}
& 
\begin{tabular}[c]{@{}l@{}}
0.2513
\end{tabular}
& 
\begin{tabular}[c]{@{}l@{}}
0.7742
\end{tabular}
& 
\begin{tabular}[c]{@{}l@{}}
0.3794
\end{tabular}
\\\midrule

\begin{tabular}[c]{@{}c@{}}
\textbf{VanillaNER}
\end{tabular}
& 
\begin{tabular}[c]{@{}l@{}}
821.67
\end{tabular}
&
\begin{tabular}[c]{@{}l@{}}
146.33
\end{tabular}
& 
\begin{tabular}[c]{@{}l@{}}
0.1781
\end{tabular}
& 
\begin{tabular}[c]{@{}l@{}}
0.7867
\end{tabular}
& 
\begin{tabular}[c]{@{}l@{}}
0.2904
\end{tabular}
\\
\bottomrule
\end{tabularx}
}
\end{center}
\caption{Quality of noise estimation. The number of true mistakes, based on our manual correction, is $186$. The potential mistakes are counted based on average of 3 runs.}
\label{tbl:noise_estimation}
\end{table*}

    \subsection{Parameter Study}
        We study how \our performs with different hyper-parameters, i.e., $t$ (the number of iterations that we run mistake estimation), $k$ (the number of folds in mistake estimation), and $\epsilon$ (the weight scaling factor of identified potential mistakes).
        
        In principle, 
        a larger $t$ usually gives us a more stable mistake estimation.
        However, a larger $t$ also requires more computation resources. In our experiments (see Table~\ref{tbl:noise_t}), we find that $t = 3$ provides a good enough result.
        
        Specifically, during mistake estimation, we have to choose the number of folds to partition the data.
        The more partitions made, the smaller each $D_i$ is and the fewer sentences will be filtered, leading to more training data $\mbox{train\_set}_i$ and better trained $M_i$. 
        On the other hand, this is at the cost of higher computational expense. 
        As shown in Table~\ref{tbl:noise_k}, we observe that $k = 5, k = 10$ are significantly better than $k = 2$. 
        In fact, when $k = 2$, each $\mbox{train\_set}_i$ has only around 5000 sentences and 1500 entities inside.
        These numbers become 7000 and 4000 when $k = 5$, and 9000 and 7000 when $k = 10$.
        
        As we mentioned before, the value $\epsilon$ can be chosen by estimating the quality of mistake estimation. 
        Table~\ref{tbl:noise_reweigh} presents some results when other values are used.
        $\epsilon = 0.3$ leads to the worst performance.
        Since our estimation does not have high precision, assigning $\epsilon$ to a low value like $0.3$ may not be a good choice.
        Interestingly $\epsilon = 0.5$ performs on par with $\epsilon = 0.7$, and even slightly better in the original test set.
        We hypothesize that this is because there are some ambiguous sentences that we did not count during estimating the quality of mistake estimation, see Section \ref{sec:case}, and the actual precision could be higher.

    \subsection{Other Datasets}
        To show the generalizability of our method across domains and languages, we further evaluate \our on an emerging-entity NER dataset from WNUT'17 and a Sinhalese NER dataset from LORELEI\footnote{LDC2018E57}.
        Sinhalese is a low-resource, morphology-rich language.
        For WNUT'17, we use the Flair as our base NER algorithm. For Sinhalese, we use BERT~\cite{devlin2018bert} followed by a BiLSTM-CRF as our base NER algorithm. 
        We use the same parameters as used in the previous CoNLL03 experiments, namely $k = 10, t = 3, \epsilon = 0.7$. 
        
\begin{table}[t]
\center
\small
\begin{tabularx}{1\linewidth}{l *{2}{Y}}
\toprule   \textbf{Dataset}        & \textbf{w/o CrossWeigh} & \textbf{w/ CrossWeigh}\\

\midrule
 WNUT'17 & 48.96 ($\pm$0.97) & 50.03 ($\pm$0.40)  \\
 \midrule
 Sinhalese & 66.34 ($\pm$0.34) & 67.68 ($\pm$0.21)  \\

\bottomrule
\end{tabularx}
\vspace{-0.15cm}
\caption{Applying CrossWeigh on other datasets}
\label{tbl:other}
\vspace{-0.3cm}
\end{table}

        The results averaged across 5 runs are reported in Table~\ref{tbl:other}. 
        One can observe quite similar results as those in the previous CoNLL03 experiments.
        Training with \our leads to a significantly higher F1 and a smaller standard deviation.
        This suggests that \our works well in other datasets and languages.

\begin{table*}[t]
\centering
\small
\scalebox{0.9}{
\begin{tabularx}{1.1\linewidth}{c ll}
\toprule
              & \multicolumn{1}{c}{\textbf{Training Set}} & 
              \multicolumn{1}{c}{\textbf{Test Set}} \\
\midrule
\textbf{Text} & Hapoel Haifa 3 Maccabi Tel Aviv 1 & Hapoel Jerusalem 0 Maccabi Tel Aviv 4 \\
\midrule
\midrule
\textbf{Original Annotations} & [Hapoel Haifa]\{ORG\}, \color{red}{\textbf{[Tel Aviv]\{ORG\}}} & [Hapoel Jerusalem]\{ORD\}, [Maccabi Tel Aviv]\{ORG\}\\
\midrule
\textbf{Correct Annotations} & [Hapoel Haifa]\{ORG\}, [Maccabi Tel Aviv]\{ORG\} & [Hapoel Jerusalem]\{ORD\}, [Maccabi Tel Aviv]\{ORG\}\\
\midrule
\midrule
 & \multicolumn{1}{c}{\textbf{Action}} & \multicolumn{1}{c}{\textbf{Result}} \\\midrule
\textbf{Flair}  &  Assumes this sentence is equally reliable as others. & [Hapoel Jerusalem]\{ORD\}, \color{red}{\textbf{[Tel Aviv]\{ORG\}}} \\
\midrule
\textbf{Flair w/ CrossWeight} & Lowers the weight of this sentence as mistakes. &  [Hapoel Jerusalem]\{ORD\}, [Maccabi Tel Aviv]\{ORG\}\\
\bottomrule
\end{tabularx}
}
\vspace{-0.15cm}
\caption{Case Study on the CoNLL03 dataset. Errors are marked with red}
\label{tbl:case_study}
\vspace{-0.4cm}
\end{table*}

\section{Case Studies} \label{sec:case}
\smallsection{Test Set Correction}
Despite the label mistakes that we have corrected, we also find some ambiguous but consistent cases.
For instances,
(1) All NBA/NHL divisions such as ``CENTRAL DIVISION'', ``WESTERN DIVISION'' were annotated as \texttt{MISC}, while all European leagues, such as ``SPANISH FIRST DIVISION'' and ``ENGLISH PREMIER LEAGUE'', are not marked as \texttt{MISC} correctly --- only ``SPANISH'' and ``ENGLISH'' are labelled as \texttt{MISC}. 
And (2) ``Team A at Team B'' is a way to say ``Team A'' as an away team playing with Team B as a home team. 
However, in almost all cases (only 1 exception out of more than 100), ``Team A'' was labelled as \texttt{ORG} while ``Team B'' was labelled as \texttt{LOC}. 
For example, in ``MINNESOTA AT MILWAUKEE'', ``NEW YORK AT CALIFORNIA'', and ``ORLANDO AT LA LAKERS'', the second sports team ``MILWAUKEE'', ``CALIFORNIA'' and ``LA LAKERS'' were always labelled as \texttt{LOC}.
Because these parts behave consistently and generally follow the annotation guideline,
we didn't touch them during the test set correction. 

\smallsection{CrossWeigh Framework}
    The mistakes in the training set can harm the generalizability of the trained model. 
    For example, in Table~\ref{tbl:case_study}, the original training sentence ``Hapoel Haifa 3 Maccabi Tel Aviv 1'' contains a label mistake, because ``Maccabi Tel Aviv'' is a sports team but was not annotated completely. 
    Interestingly, there is a similar sentence in the test set -- ``Hapoel Jerusalem 0 Maccabi Tel Aviv 4''.
    In all 5 different runs of the original Flair model, they failed to predict correctly that ``Maccabi Tel Aviv'' in the test sentence as \texttt{ORG} because of the label mistake in the training sentence, even though ``\texttt{ORG} number \texttt{ORG} number'' is an obvious pattern in the training set.
    In \our, this label mistake in the training set was detected in all $t=3$ iterations and therefore assigned a very low weight during training. 
    After that, in all 5 different runs of Flair w/ \our, they successfully predict that ``Maccabi Tel Aviv'' is \texttt{ORG} as a whole.


\section{Related Work}\label{sec:rel}
In this section, we review related works from three aspects, mistake identification, cross validation \& boosting, and NER algorithms.
\subsection{Mistake Identification}
    Researchers have noticed the label mistakes in sophisticated natural language processing tasks for a while.
    For example, it is reported that the inter-annotator agreement is about 97\% on the Penn Treebank POS tagging dataset~\cite{manning2011part, subramanya2010efficient}.
    
    There are a few attempts towards detecting label mistakes automatically.
    For example, \citet{nakagawa2002detecting} designed a support vector machine-based model to assign weights to examples that were hard to classify in the POS tagging task.
    \citet{loftsson2009correcting} further applied previous detection models and manually corrected Icelandic Frequency Dictionary~\cite{pind1991islensk} POS tagging dataset.
    However, these two methods are specifically developed for POS tagging and cannot be directly applied to NER. 
    
    Recently, \citet{rehbein2017detecting} extends variational inference with active learning to detect label mistakes in ``silver standard'' data generated by machines.
    In this paper, we focus on detecting label mistakes in ``gold standard'' data, which is a different scenario.



\subsection{Cross Validation \& Boosting}
    Our mistake estimation module shares some similarity with cross validation. Applying cross validation to the training set is the same as our mistake estimation module, except that we have an \textit{entity disjoint filtering} step. Experiments in Table~\ref{tbl:entity_disjoint} show that this step is crucial to our performance gain. The choice of ten folds also stems from cross validation~\cite{kohavi1995study}.
    
    Another similar thread of work is boosting, such as Adaboost~\cite{freund1999short, schapire1999improved}. 
    For example, \citet{abney1999boosting} has applied Adaboost on the Penn Treebank POS tagging dataset and gained encouraging results on model performance.
    In boosting algorithms, the training data is assumed to be perfect.
    Therefore, it trains models using the full training set and then increases the weights of training instances that fails the current model in the next round of learning.
    In contrast, we decrease the weights of sentences that differ from the model built upon the entity disjoint training set.
    More importantly, our framework is a better fit for neural models, because they can likely overfit the training data and thus being bad choices as weak classifiers in boosting.

\subsection{NER Algorithms}
Neural models have been widely used for Named Entity Recognition, and the state-of-the-art models integrate LSTMs, conditional random field and language models~\cite{lample2016neural, ma2016end, liu2018empower, peters2018deep, akbik2018contextual}. 
In this paper, we focus on improving the annotation quality for NER, and our method has a big potential to help other methods, especially for noisy datasets~\cite{shang2018learning}. 

\section{Conclusion \& Future work}\label{sec:con}


In this paper, we explore and correct the label mistakes in the CoNLL03 NER dataset. 
Based on the corrected test set, we re-evaluate most of recent NER models.
We further propose a novel framework, \our, that is able to detect label mistakes in the training set and then train a more robust NER model accordingly.
Extensive experiments demonstrate the effectiveness of \our on three datasets and also indicate the potentials of using \our to improve the annotation quality during the label curation process.

In future, we plan to extend our framework into an iterative setting, similar to those boosting algorithms.
The bottleneck of doing this lies in the efficiency problems of training multiple deep neural models hundreds of times.
One solution to overcome it is to apply meta learning. 
We can first train a meta model and only fine-tune on different training data on each fold.
In this way, we can identify label mistakes more accurately and obtain a series of weighted models at the end.


\section*{Acknowledge}
We thank all reviewers for valuable comments and suggestions that brought improvements to our final version.
Research was sponsored in part by U.S. Army Research Lab. under Cooperative Agreement No. W911NF-09-2-0053 (NSCTA), DARPA under Agreement No. W911NF-17-C-0099, National Science Foundation IIS 16-18481, IIS 17-04532, and IIS-17-41317, DTRA HDTRA11810026,  Google Ph.D. Fellowship and grant 1U54GM114838 awarded by NIGMS through funds provided by the trans-NIH Big Data to Knowledge (BD2K) initiative (www.bd2k.nih.gov). Any opinions, findings, and conclusions or recommendations expressed in this document are those of the author(s) and should not be interpreted as the views of any U.S. Government. The U.S. Government is authorized to reproduce and distribute reprints for Government purposes notwithstanding any copyright notation hereon.
This research was supported by grant 1U54GM114838 awarded by NIGMS through funds provided by the trans-NIH Big Data to Knowledge (BD2K) initiative (www.bd2k.nih.gov). This work was supported by Contracts HR0011-15-C-0113 and HR0011-18-2-0052 with the US Defense Advanced Research Projects Agency (DARPA).

\bibliography{citation.bib}
\bibliographystyle{acl_natbib}

\end{document}